\pdfoutput=1
\documentclass[11pt]{article}

\usepackage{EACL2023}

\usepackage{times}
\usepackage{latexsym}
\usepackage[colorinlistoftodos]{todonotes}

\usepackage{amssymb}
\usepackage{multirow}
\usepackage{graphics}
\usepackage{tabularx}
\usepackage{enumitem}
\usepackage[T1]{fontenc}
\usepackage[utf8]{inputenc}
\usepackage{graphicx}
\graphicspath{ {./images/} }
\usepackage{microtype}

\usepackage{inconsolata}
\usepackage{makecell}

\usepackage{comment}

\title{Natural Language Processing in Ethiopian Languages: Current State, Challenges, and Opportunities
}
\author{\normalsize Atnafu Lambebo Tonja $^{1,*}$, Tadesse Destaw Belay$^{2,*}$, Israel Abebe Azime$^{3,*}$,\\ 
\textbf{\normalsize Abinew Ali Ayele$^{4,5,*}$,  Moges Ahmed Mehamed$^{6,*}$, Olga Kolesnikova$^{1}$, Seid Muhie Yimam$^{5,*}$,}\\
\footnotesize
$^*$EthioNLP, $^1$Instituto Politécnico Nacional, Mexico, $^2$Wollo University, Ethiopia, $^3$ Saarland University, Germany,
  \\
 \footnotesize
$^4$Bahir Dar University, Ethiopia,$^5$Universität Hamburg, Germany, $^6$Wuhan University of Technology, China.
 \\
}

\begin{document}
\maketitle

\begin{abstract}
% This survey paper will discuss the state-of-the-art Natural Language Processing (NLP) and resources for Ethiopian languages. 
%\todo[color=blue]{this needs summary}

This survey delves into the current state of natural language processing (NLP) for four Ethiopian languages: Amharic, Afaan Oromo, Tigrinya, and Wolaytta. Through this paper, we identify key challenges and opportunities for NLP research in Ethiopia.
Furthermore, we provide a centralized repository on GitHub %\footnote{\url{https://github.com/EthioNLP/survey}}
  that contains publicly available resources for various NLP tasks in these languages. This repository can be updated periodically with contributions from other researchers.
Our objective is to identify research gaps and disseminate the information to NLP researchers interested in Ethiopian languages and encourage future research in this domain.

\end{abstract}
\section{Introduction}
% Humans utilize language as a form of communication for a variety of tasks in their daily lives \cite{sorensen1949hf}. We use communication in general to convey information about the outside world, but we also use it to give instructions, ask questions, and express sentiments. 
Due to the rise of its applications in many fields, Natural Language Processing (NLP), a sub-field of Artificial Intelligence (AI), is receiving a lot of attention in terms of research and development \cite{kalyanathaya2019advances}.
% The tools and processes developed by many researchers who worked on NLP are what make it what it is today.
NLP tasks such as Machine Translation (MT), Sentiment or Opinion Analysis, Parts of Speech (POS) Tagging, Question Classification (QC) and Answering (QA), Chunking, Named Entity Recognition (NER), Emotion Detection, and Semantic Role Labeling is currently highly researched areas in different high-resource languages.% because of their significant contributions to NLP.

Because of the advancement of deep learning and transformer approaches, modern NLP systems rely largely on the availability of vast volumes of annotated and unannotated data to function well. The majority of the languages in the world do not have access to such enormous information tools, despite the fact that a few high-resource languages have received more attention. 
\begin{figure}[h!]
    \centering
    \includegraphics[width=\linewidth]{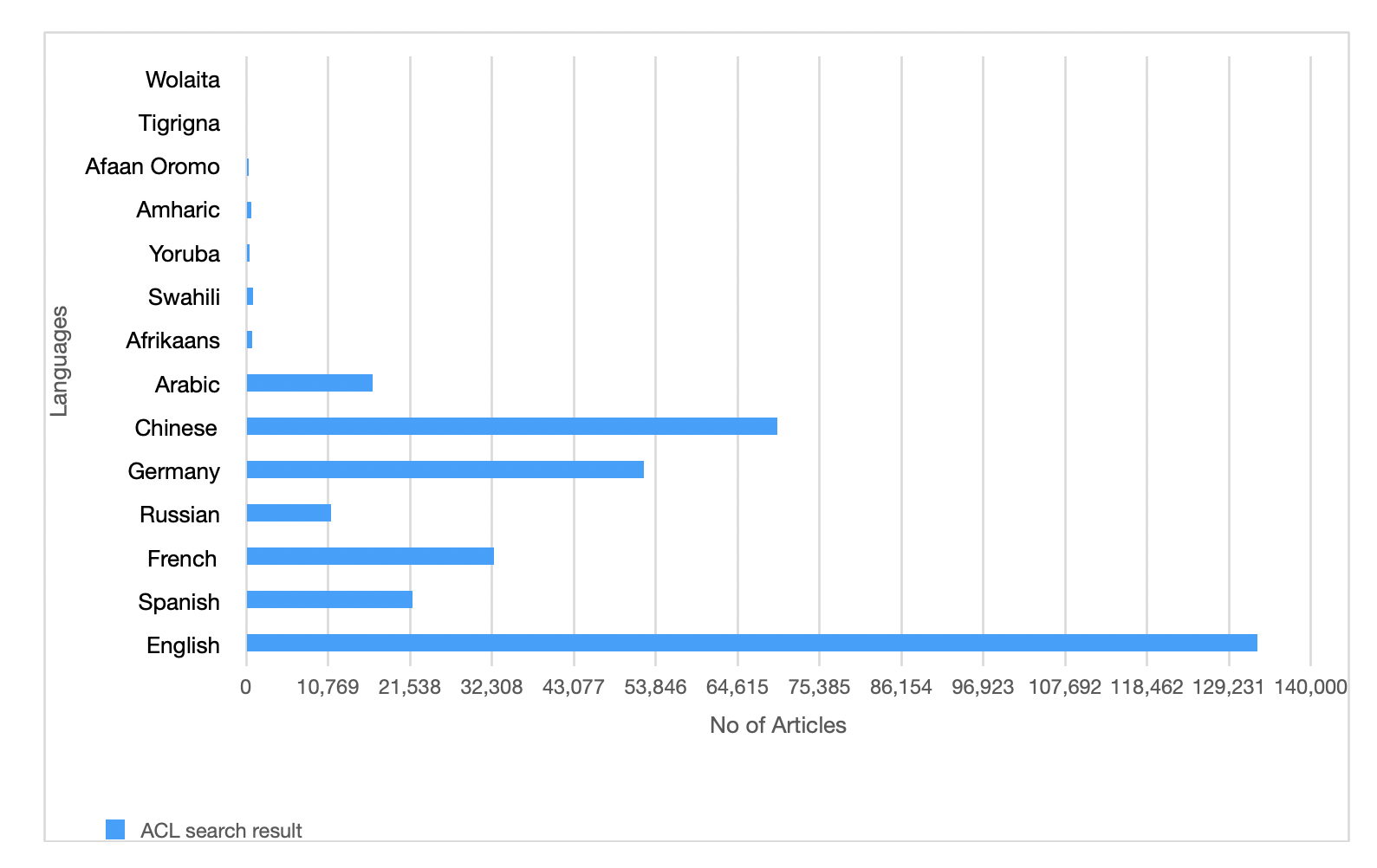}
    \caption{ACL paper search results for high and low-resource languages. }
    \label{fig:aclsearch}
\end{figure}
Ethiopia is a country with more than 85 spoken languages, but only a few are presented in NLP progress. Figure \ref{fig:aclsearch} shows a search result for articles found in the ACL anthology for high and low-resource languages. As we can see from Figure \ref{fig:aclsearch}, the search result for low-resource languages shows a very insignificant number of research works, while high-resource languages like English dominate in the ACL anthology paper repository. This might be a reflection of the unavailability of resources in the digital world, like in other high-resource languages, which affected the NLP progress in low-resource languages in general and Ethiopian languages in particular.

In this paper, we overview research works done in the area of selected NLP tasks for four Ethiopian languages. We cover mainly the following 4 languages, namely Amharic (Amh), Afaan Oromo (Orm), Tigrinya (Tir), and Wolaytta (Wal). We also reviewed works on a small set of local languages including Awigna (Awn) and Kistane(Gru), specially for the machine translation tasks. 
% We discuss the NLP progress, challenges, and opportunities and point to potential directions for future work.
% This section will state selected Ethiopian Language Characteristics, targeted NLP tasks, motivations, and  contributions of the survey.
The contributions of this paper are as follows:
% While tools and resources are readily available to produce semantic features for high-resource languages like Arabic, Chinese, English, or German, it is much harder to obtain any semantically-biased information for low-resource languages such as Ethiopian languages.
%\begin{itemize}
    \noindent \textbf{(1)} Reporting the current state-of-the-art NLP research for Ethiopian languages. 
    % who are new to the field
    \noindent \textbf{(2)} Discussing NLP progress in Ethiopian languages and the main challenges and opportunities for Ethiopian NLP research.
    % \item We discuss different recommendations for the new researchers and people interested to work in Ethiopian languages 
    \noindent \textbf{(3)} Collecting and presenting publicly available resources for different NLP tasks in Ethiopian languages in one GitHub repository that can be extended periodically in collaboration with other researchers.
    % central place
%\end{itemize}
The collected publicly available datasets and models for Ethiopian languages are in our GitHub repository\footnote{https://github.com/EthioNLP/Ethiopian-Language-Survey}.
% The reset the paper is organized as follows :\todo[color =blue]{will be added}

\section{Language Details} 
This paper assesses the progress of NLP research for four Ethiopian languages: Amharic, Afaan Oromo, Tigrinya, and Wolaytta. As Ethiopia is a multilingual, multicultural, and multi-ethnic country, those selected languages have more speakers and native speakers in the country. Additionally, we have searched papers in the major eight Ethiopian languages and taken the top four based on frequency from the ACL anthology. This section gives some descriptions of those four targeted languages. 

\textbf{Amharic}: is an Ethio-Semitic and Afro-Asiatic language. It is the official working language of the Federal Democratic Republic of Ethiopia (FDRE). It has about 57 million speakers, which makes it the second most spoken Semitic language in the world,  where 32 million of them are native speakers \citep{Eberhard2022}. Other known names for this language are Amarigna and Amarinya.

\textbf{Afaan Oromo}: is a Cushitic language family. The language name may be written in different alternatives: (Afan, Afaan, affan) Oromo, or simply Oromo. There are over 50 million native Oromo speakers \citep{Eberhard2022}.

\textbf{Tigrinya}: (alternatively: Tigregna, Tigrinya or Tigrigna) is a Semitic language family spoken in the Tigray region of Ethiopia and in Eritrea. The language uses Ge'ez script with some additional alphabets that have been created for the Tigrinya language and are closely related to Ge'ez and Amharic. The language has around 9.9 million native speakers \citep{Eberhard2022}.

\textbf{Wolaytta}: (alternatively: Wolayita, Wolaytegna, Wolaytigna, Welaitta, and Welayita) is an Omotic language family spoken in  the Wolaytta zone of Ethiopia. Both Afan Oromo and Wolaytta are written in the Latin alphabet. 

\begin{comment}
The languages and the ISO code abbreviations that are included in this paper are shown in Table \ref{tab:lang}.
     
\begin{table}[h!]
\small
\centering
\begin{tabular}{llll}
\hline
\textbf{Language} & \textbf{ISO code} & \textbf{Spoken in} & \textbf{Script}\\
\hline
Amharic & Amh & Ethiopia & Ethiopic \\
English & Eng & World & Latin \\
Oromo & Orm & Ethiopia & Latin \\
Tigriyna & Tir & Ethiopia, Eritrea & Ethiopic \\
Wolaytta & Wal & Ethiopia & Latin \\
Kistane & Gru & Ethiopia & Ethiopic\\
Awngi & Awn & Ethiopia & Ethiopic \\
\hline
\end{tabular}
\caption{\label{tab:lang}
Language abbreviations and descriptions are used in this work.}
\end{table}
\end{comment}

\section{Low-resource Languages} \label{low}
Researchers concerned with NLP have used data availability (either in the form of labeled, unlabeled, or auxiliary data) and NLP tools and resources as criteria for defining low-resource languages \cite{ranathunga2021neural}. According to the work by \citet{gereme2021combating}, low-resource languages lack the tools and resources important for NLP and other techno-linguistic solutions. In addition, low-resource languages lack new language technology designs. Due to all these limitations, it is very difficult to develop new powerful methods for language applications \cite{tonja2023low}. For resource-rich languages such as English, Germany, French, Spanish and etc, the size of the dataset is not a problem because researchers have created a large set of corpora and tools for many NLP tasks. However, many other languages are deemed to be low-resource languages \cite{fesseha2021text}. With this intuition, Ethiopian languages such as Amharic \cite{gereme2021combating}, Afaan Oromo \cite{abate2019english}, Tigrinya \cite{osman2012stemming}, Wolayitta \cite{tonja2023low} are "low-resource" languages due to lack of data resources, linguistic materials, and tools. This affected the development of different NLP tasks and tools.

\section{Possible Resource Sources and Tools}
Data is one of the crucial building blocks for any NLP application. The availability of data is one of the criteria to categorize one language as a high or low-resource language \cite{ranathunga2021neural}. As discussed in Section \ref{low} Ethiopian languages belong to low-resource languages due to the unavailability of data. Table \ref{datasource} shows some possible digital data sources for different NLP tasks.  

\begin{table*}[h!]
% \resizebox{\textwidth}{!}{%
\begin{tabular}{l|c|l}
\hline
\multicolumn{2}{c|}{\textbf{Sources}} & \textbf{Link} \\ \hline
\multirow{2}{*}{Religion books} 
& Bible & \url{https://www.bible.com/} \\ \cline{2-3} 
& Quran &   \\ \hline
\multirow{4}{*}{Multilingual data repositories} 
& Opus         & \url{https://opus.nlpl.eu     }               \\ \cline{2-3} 
& Lanfrica     & \url{https://lanfrica.com }                   \\ \cline{2-3} 
& Masakhane    & \url{https://github.com/masakhane-io }         \\ \cline{2-3} 
& Hugging face & \url{https://huggingface.co/    }              \\ \hline
\multirow{5}{*}{News medias}                  
& Fana         & \url{https://www.fanabc.com }                 \\ \cline{2-3} 
& EBC          & \url{https://www.ebc.et}                      \\ \cline{2-3} 
& BBC          & \url{https://www.bbc.com }                    \\ \cline{2-3} 
& DW           & \url{https://www.dw.com} \\ \cline{2-3} 
& Walata       & \url{https://waltainfo.com/  }                 \\ \hline
\multirow{3}{*}{Social medias}                
& Twitter      & \url{https://twitter.com/}                     \\ \cline{2-3} 
& Facebook     & \url{https://www.facebook.com/ }               \\ \cline{2-3} 
& Reddit       & \url{https://www.reddit.com/ }               \\ \hline
\multirow{1}{*}{Text Corpus} 
& Amharic Text Corpus & \href{https://data.mendeley.com/datasets/dtywyf3sth/1}{Amharic Corpus at Mendeley}\\
\hline
\end{tabular}
% }
\caption{Possible data sources \label{datasource}}
\end{table*}

%\section{NLP Tools} \label{NLPtool}
Like data, NLP tools are also one of the building blocks for NLP applications, and the unavailability of these tools for a certain language also directly affects the development of NLP applications for that language. Table \ref{tab:tool} shows available NLP tools for Ethiopian languages. As it can be seen from Tables \ref{datasource} and \ref{tab:tool}, there are still very few sources to gather digital data and tools, available for Ethiopian languages.

% \todo{Lets cover the basic tools such as Morphological analysis (example HornMorpho), segmenter and tokenizer (example amseg), transliteration and normalizer (example amseg). Let's create a table as well, that indicates at least availability}

\begin{table*}[h!]
\resizebox{\textwidth}{!}{%
\centering
\begin{tabular}{l|l|l|l|l}
\hline
\textbf{Author (s)} & \textbf{Tool's name} & \textbf{Tool's task} & \textbf{Language (s) support} & \textbf{Resource link}\\
\hline
\citet{yimam2021introducing,belay-etal-2021} & amseg & \makecell{Segmenter, tokenizer, transliteration,\\romanization and normalization} & Amh & \href{https://pypi.org/project/amseg/}{amseg} \\

\citet{gasser2011hornmorpho} & HornMorpho & Morphological analysis & Amh, Orm, Tig & \href{https://github.com/hltdi/HornMorpho}{HornMorpho} \\

\citet{seyoum2018universal} & lemma & Lemmatizer & Amh & \href{https://universaldependencies.org/}{Lemmatizer} \\
\hline
\end{tabular}
}
\caption{\label{tab:tool}
Available language tools that are developed for low-resource Ethiopian languages.}
\end{table*}

\section{NLP Tasks and Their Progress} \label{NLP}
In this section, we discuss what work has been done, what datasets of what sizes were used, what methods or approaches the authors proposed, and the availability of their dataset and models for NLP tasks and their progress in selected Ethiopian languages. We focused on Machine Translation (MT), Part-of-speech (POS) tagging, Named Entity Recognition (NER), Question Classification (QC), Question Answering (QA), text classification, and text summarization tasks due to the large number of works done for the targeted low-resource languages. The available models, the datasets for the tasks, and their links are found in Table \ref{related}. 

\subsection{POS Tagging}
POS tagging is one of the popular NLP tasks that refer to categorizing words in a text (corpus) in correspondence with a particular part of speech, depending on the definition of the word and its context \cite{pailai2013comparative}. 
% This section discusses POS tagging work done for four Ethiopian languages.
\begin{table*}[ht]
%\tiny
\centering
 \resizebox{\textwidth}{!}{%
\begin{tabular}{c|lrrrll}
%\toprule
\hline
 \textbf{Languages}&\textbf{Author(s)}  &\textbf{Size} & \textbf{Approach} &\textbf{Score} & \textbf{Dataset}  & \textbf{Model}   \\ \hline
 \multirow{7}{*}{Amharic}&\citet{adafre2005part}  & 1000 &CRF&74.00 &No&No  \\ \cline{2-7}
 &\citet{gamback2009methods} & 210,000 &MaxEnt&94.52 &No&No  \\ \cline{2-7}
 &\citet{tachbelie2009amharic} & 210,000 &SVM&85.50 &No&No  \\ \cline{2-7}
 &\citet{gebre2010part} & 206,929 &SVM&90.95 &No&No  \\ \cline{2-7}
  &\citet{hirpssa2020pos} & 210,000 &CRF&94.08 &No&No\\ \cline{2-7}
  &\citet{yimam2021introducing} & 210,000 &CRF&92.27 &Yes&Yes\\ \cline{2-7}
    &\citet{gashaw2020machine} & 109,676 &CRFSuit&95.10 &No&No\\ \cline{2-7}
  &\citet{tachbelie2011part} & 210,000  &MBT&93.51 &No&No    \\ \hline
  \multirow{2}{*}{Afaan Oromo}&\citet{wegari2011parts}  & 1621  &HMM&91.97 &No&No  \\ \cline{2-7}
 &\citet{ayana2015towards} & 17,473 & Bill’s tagger&95.60 &No&No  \\ \hline
  \multirow{2}{*}{Tigrinya}&\citet{tedla2016tigrinya}  & 72,080  &CRF &90.89&Yes&No  \\ \cline{2-7}
 &\citet{tesfagergish2020deep} & 72,080  & LSTM&91.00 &No&No  \\ \hline
 \multirow{1}{*}{Wolayitta}&\citet{shirko2020part}  & 14,358 &HMM&92.96  &Yes&No  \\ \hline
\end{tabular}
}
\caption{Summary of related works for selected Ethiopian languages in POS tag tasks, \textbf{Size} shows the number of tokens used during the experiment, \textbf{Score} shows the outperformed model results evaluated using accuracy score, \textbf{Dataset} and\textbf{ Model} shows the availability of dataset and models in publicly accessible repositories.\label{tab:possum}}
\end{table*}

Table \ref{tab:possum} summarizes the current state of POS tagging research for selected Ethiopian languages. The table shows the name(s) of the author(s), the size of the dataset, the method used, the accuracy score of the models, and the availability of datasets and models in public repositories.

For Amharic, seven studies are listed, which used different approaches such as Conditional Random Fields (CRF), Maximum Entropy (MaxEnt), Support Vector Machines (SVM), CRFSuit, and Memory-Based Tagger (MBT). The highest accuracy score was achieved using the CRFSuit approach by \citet{gashaw2020machine}. For Afaan Oromo, two studies are listed that used the Hidden Markov Model (HMM) and  Brill's tagger. The highest accuracy score was achieved using Bill's tagger by \citet{ayana2015towards}. For Tigrinya, two studies are listed that used CRF and Long Short-Term Memory (LSTM). The highest accuracy score was achieved by 
the LSTM approach in \citet{tesfagergish2020deep} and for Wolayitta, one study is listed, that used HMM and achieved an accuracy score of 92.96. 

From Table \ref{tab:possum}, we can conclude that POS tagging is less researched for Ethiopian languages, the majority of the works were found for Amharic than for the other languages. From the works discussed in Table \ref{tab:possum} only the work by \citet{yimam2021introducing} made their models and datasets available for public use. 

\subsection{Named Entity Recognition (NER)}
In this section, we present works related to Named Entity Recognition (NER) for Ethiopian languages.

For \textbf{Amharic}, \citet{mehamed2019named} conducted the NER experiment on a corpus of 10,405 tokens using the CRF classifier.  \citet{alemu2013named} conducted the experiments on a manually developed corpus of 13,538 words with the Stanford tagging scheme. \citet{tadele2014amharic} used a hybrid of machine learning (decision trees and support vector machines) and rule-based methods. The datasets for these works are not available. 
The work done by the Masakhane NLP group \cite{tacl_a_00416} analyzed a 10 African languages dataset and conducted an extensive empirical evaluation of state-of-the-art methods across both supervised and transfer learning settings, including Amharic. The data and models are available on GitHub. \citet{gamback2017named,yimam2021introducing,sikdar2018named} built a deep learning-based NER system for Amharic using the available SAY project NER dataset. \citet{jibril2022anec} proposed a transformer-based NER recognizer for Amharic using a new annotated 182,691 word dataset. All available NER datasets for Ethiopian languages are shown in Table \ref{related}.

For \textbf{Afan Oromo}, the work by \citet{legesse2012named} implemented the first NER system using a hybrid approach (rule-based and statistical) which contains 23k words. \citet{abdi2015afaan} deals with NER in a hybrid (machine learning and rule-based) approach using the data from the work of \citet{legesse2012named}. \citet{abafogi2021boosting} adopted boosting NER by combinations of such approaches as, machine learning, stored rules, and pattern matching using 44k words out of which around 7.8k were named entities.    
\citet{gardie2022afan} developed a NER system using 12,479 data instances and BiLSTM,  word embedding, and CNN approaches.  However, none of the datasets in the above Afan Oromo works are publicly available.  

For \textbf{Tigrinya}, the research by \citet{yohannes2022named} proposed a method for NER using a pre-trained language model, TigRoBERTa. The dataset contains 69,309 manually annotated words. Later \citet{yohannes2022method} employed Tigrinya NER with an addition of 40,627 words.

\begin{figure}[h!]
\includegraphics[width=\linewidth]{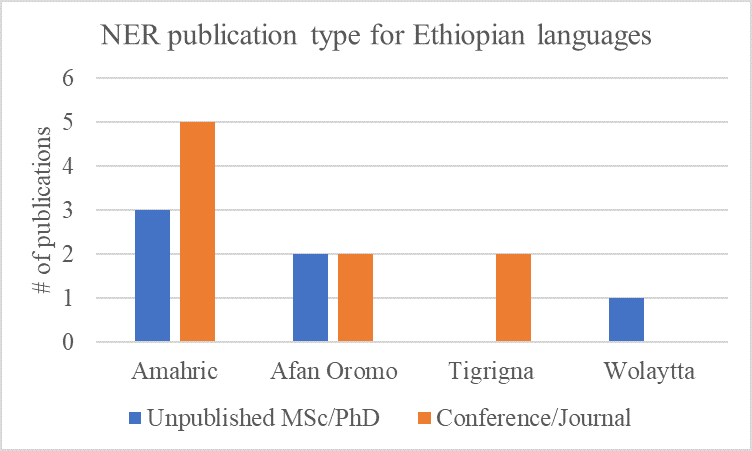} %width=8cm
\centering
\caption{NER publication types for Ethiopian languages: the figure description is the same as in Figure \ref{fig:qa_types}. Wolaytta has no published works in conferences/journals.}
\label{fig:ner_types}
\end{figure}

\begin{figure}[h!]
\includegraphics[width=\linewidth]{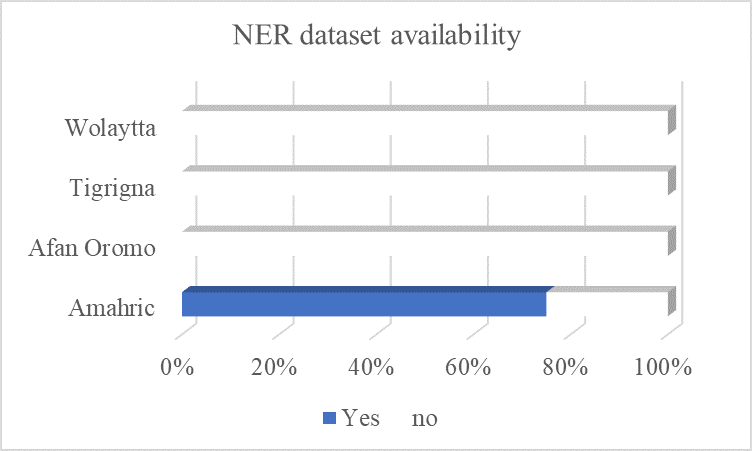}
\centering
\caption{NER dataset availability per language: as it can be seen relatively more NER datasets are available only for the Amharic language.}
\label{fig:ner_dataset}
\end{figure}

The only NER work attempted \textbf{Wolaytta} Language was conducted by \citet{biruk2021named} using a machine learning approach. Figures \ref{fig:ner_types} and \ref{fig:ner_dataset} show NER publication types and  dataset availability for targeted Ethiopian languages, respectively.
%
%NER dataset availability
We can summarize that NER is a little more developed than the POS tagging for Afan Oromo, Tigriyna, and Wolaytta languages. However, like POS tagging, only small Amharic NER datasets shown in Figure \ref{fig:ner_dataset} are available. 

\subsection{Machine Translation (MT)} \label{MT}
% As computational activities become more mainstream and the internet opens up to a wider multilingual and global community, research and development in MT continue to grow at a rapid rate \cite{kenny2018machine}. High translation
% quality has been achieved for high-resource language pairings such as English-Spanish, English-
% French, English-Russian, and English-Portuguese. MT systems perform poorly in
% low-resource environments due to a lack of enough training data for those languages. 
With the increasing popularity of computational tasks and the Internet's expanding reach to diverse, multilingual communities, the field of MT is rapidly progressing \cite{kenny2018machine}. While impressive translation results have been achieved for language pairs with abundant resources, such as English-Spanish, English-French, English-Russian, and English-Portuguese, MT systems struggle in environments with limited resources, where insufficient training data for certain languages is the main obstacle.
In this section, we discuss the MT progress for Ethiopian languages in three categories: \textbf{(i) \textit{English Centeric-}} works done for the above target Ethiopian languages with English pair, \textbf{(ii) \textit{Ethiopian - Ethiopian -}} works done for Ethiopian language pairs without involving other languages,  and  \textbf{(iii) \textit{Multilingual MT -}} works done for Ethiopian languages with other languages in a multilingual setting.  
% Table \ref{tab:mtsum} depicts the summary of related works in MT for selected Ethiopian languages.

\begin{table*}[h!]
%\tiny
\centering
 \resizebox{\textwidth}{!}{%
\begin{tabular}{c|llrlrll}
%\toprule
\hline
\textbf{Categories} & \textbf{Author(s)} & \textbf{Lang. pairs} &\textbf{Size} & \textbf{Approach} &\textbf{Score} & \textbf{Dataset}  & \textbf{Model}   \\ \hline
\multirow{5}{*}{English centeric} &  \citet{biadgligne2021parallel} & Amh-Eng & 231,898  &NMT&32.44 &No&No  \\ \cline{2-8} 
                                  & \citet{gezmu2021extended} & Amh-Eng & 145,364 & NMT &32.20&Yes&No\\ \cline{2-8} 
                                  & \citet{ashengo2021context} & Amh-Eng & 8,603  &RNNMT&21.46 &No&No \\ \cline{2-8} 
                                  &  \citet{biadgligne2022offline} & Amh-Eng & 231,898 & NMT&37.79 &No&No \\ \cline{2-8} 
                                  & \citet{belay2022effect} & Amh-Eng & 1,140,130 &NMT &37.79 &No&No  \\ \cline{2-8}
                                  & \citet{solomon2017optimal} & Orm-Eng &6,400  &SMT & 47.00 &No &No  \\ \cline{2-8}
                                  & \citet{meshesha2018english} & Orm-Eng & 6,400 &SMT&27.00  &No &No \\ \cline{2-8}
                                  & \citet{adugna2010english} & Orm-Eng & 21,085 &SMT&  17.74 &No &No \\ \cline{2-8}
                                  & \citet{chala2021crowdsourcing} & Orm-Eng & 40,000  &NMT& 26.00 &No &No \\ \cline{2-8}
                                  &\citet{gemechu2021machine} & Orm-Eng & 10,000    &NMT &41.62  & No &No\\ \cline{2-8}
                                  &\citet{tedla2016effect} & Tir-Eng & 31,279    &SMT &20.90  & No&No\\ \cline{2-8}
                                  &\citet{tedla2017morphological} & Tir-Eng & 31,279    &SMT & 20.00  & No &No\\ \cline{2-8}
                                  &\citet{berihu2020enhancing} & Tir-Eng &32,000    &Hybrid &67.57  & No &No\\ \cline{2-8}
                                  &\citet{azath2020statistical} & Tir-Eng &  17,338    &SMT & 23.27 & No &No\\ \cline{2-8}
                                  &\citet{kidane2021exploration} & Tir-Eng  & 336,000 &NMT& 15.52 & Yes &No\\ \cline{2-8}
                                  &\citet{tonja2021parallel} & Wal-Eng  & 26,943 & NMT& 13.80 & No &No\\ \cline{2-8}
                                  &\citet{tonja2023low} & Wal-Eng  & 26,943 &NMT & 16.10  & No &No\\ \cline{2-8}
                                  &\citet{abate2019english} & Amh-Eng &40,726 &SMT & 13.31 & Yes & No \\ \cline{2-8}
                                 & \citet{abate2019english}&Orm-Eng &14,706 & SMT & 14.68 & Yes & No\\ \cline{2-8}
                                &\citet{abate2019english}& Tir-Eng&35,378&SMT& 17.89& Yes&No\\\cline{2-8}
                                &\citet{abate2019english}&Wal-Eng & 30,232 &SMT &10.49 &Yes &No\\ \cline{2-8}
                                  \hline
\multirow{3}{*}{Local -Local} &  \citet{Mekonnen_2019} & Amh-Awn & 5,000  &SMT& 17.26 &No &No  \\ \cline{2-8} 
                                  & \citet{10.1007/978-3-319-95153-9_13} & Amh-Tir & 27,000 &SMT&9.11 &No &No\\ \cline{2-8} 
                                  & \citet{ashengo2021context} & Amh-Gur & 9,225  &NMT& 7.73&No&No\\ \hline
\multirow{5}{*}{Multilingual} &  \citet{lakew2020low} & Amh-Eng & 373,358  &NMT & 20.86 &  Yes   
                                  &Yes \\ \cline{2-8} 
                                  &\citet{lakew2020low} & Orm-Eng & 14,706&NMT  & 32.24&  Yes   
                                  &Yes \\ \cline{2-8} 
                                 &  \citet{lakew2020low} & Tir-Eng & 917,632&NMT&32.21   &  Yes   
                                  &Yes  \\ \cline{2-8} 
                                  & \citet{vegi-etal-2022-webcrawl} & Amh-Eng & 46,000 & NMT &24.17 &Yes&No\\ \cline{2-8} 
                                  & \citet{vegi-etal-2022-webcrawl} & Orm-Eng & 7,000 &NMT & 12.13&Yes&No\\ \hline
\end{tabular}
}
\caption{Summary of related works for selected Ethiopian languages in MT task, \textbf{Lang. pairs} is language pairs used for translation, \textbf{Size} shows the number of parallel sentences used in each paper, \textbf{Score} shows the outperformed model results evaluated using BLEU score, \textbf{Dataset} and\textbf{ Model} shows the availability of dataset and models in publicly accessible repositories, respectively.\label{tab:mtsum}}
\end{table*}

 Table \ref{tab:mtsum} summarizes several studies on MT in selected Ethiopian languages, focusing on the three categories. The studies vary in size of the parallel dataset, approach, score, availability of dataset, and model for public use. For English-centric language pairs, five studies used Amh-Eng language pairs. \citet{biadgligne2021parallel} used NMT, and the size of their dataset was 231,898, while \citet{gezmu2021extended} used NMT and had a dataset size of 145,364. \citet{ashengo2021context} used RNNMT, and their dataset size was 8,603. The study by \citet{biadgligne2022offline} used NMT, and their dataset size was the same as that of \citet{biadgligne2021parallel}. \citet{belay2022effect} used NMT with a dataset size of 1,140,130. Finally, four studies used language pairs: Orm-Eng, Tir-Eng, and Wol-Eng. These studies applied different approaches such as SMT, NMT, and hybrid, with dataset sizes ranging from 6,400 to 336,000.

Three studies used Amharic (Amh) and other local language pairs, with different approaches and parallel dataset sizes. The study by \citet{Mekonnen_2019} used Amh-Awn language pairs, with a parallel dataset size of 5,000 and an SMT approach, while the study by \citet{10.1007/978-3-319-95153-9_13} worked on Amh-Tir language pairs with a parallel dataset size of 27,000 and an SMT approach. Finally, the study by \citet{ashengo2021context} used Amh-Gur language pairs with a parallel dataset size of 9,225 and an NMT approach. The performance in these studies ranged from 7.73 to 17.26 in BLEU scores. For multilingual MT, we found two studies by \citet{lakew2020low} and \citet{vegi-etal-2022-webcrawl} that included Ethiopian languages with other African languages. 

In Table \ref{tab:mtsum}, we can see some of the notable findings of the studies, for example, \citet{solomon2017optimal} achieved a high BLEU score of 47.00 with their SMT approach, although their parallel dataset size was small (6,400). The study by \citet{berihu2020enhancing} used a hybrid approach and achieved a high BLEU score of 67.57 with a parallel dataset size of 32,000. \citet{kidane2021exploration} used NMT with a large parallel dataset size (336,000), but their BLEU score was relatively low (15.52). \citet{tonja2021parallel} and \citet{tonja2023low} used NMT for Wal-Eng language pairs, with parallel dataset sizes of 26,943, but their scores were relatively low (13.8 and 16.1, respectively). Lastly, in multilingual MT studies, the work by \citet{lakew2020low} made the datasets and models available for public use. More analysis of MT studies for the selected languages are discussed in Appendix \ref{appMT}.

\subsection{Question Answering and Classification}
Even though question classification (QC) and question answering (QA) have been largely studied for various languages, they have barely been studied for Amharic, Afaan Oromo, Tigrinya, and Wolaytta. Some of the QC and QA work conducted for these languages are discussed below.

For \textbf{Amharic}, the work by \citet{saron2021} implemented a Convolutional Neural Network (CNN) based Amharic QC model using around 8k generic Amharic questions from different websites and labeled into 6 classes, similar to the question classes proposed by \citet{li2006learning}. The work done by \citet{libsie-2019-amharic} developed Amharic non-factoid QA for biography, definition, and description questions. \citet{yimam2009teteyeq} developed an Amharic QA system for factoid questions. However, the datasets of the aforementioned works are still not available for further investigation. \citet{nega2016question} presented machine learning (SVM) based Amharic QC using a total of 180 questions collected from the Agriculture domain. Lastly, the work done by \citet{belay-etal-2021} built a QC dataset from a Telegram public channel called \href{https://t.me/askAnythingEthiopia}{Ask Anything Ethiopia} and developed deep learning-based Amharic question classifiers. \citet{nega2016question} and \citet{belay-etal-2021} datasets are released in a GitHub repository (see Table \ref{related}). 

%%%%%%%%%%%%%%
% paused here for github
%%%%%%%%%%%%%%%

\begin{figure}[h!]
\includegraphics[width=\linewidth]{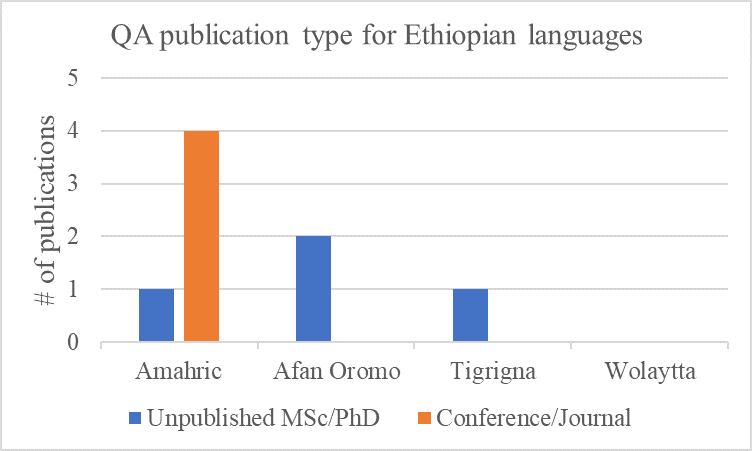} %width=8cm
\centering
\caption{QC/QA publication type: MSc/Ph.D. is an unpublished master or Ph.D. thesis uploaded in local universities repositories and archives. A Conference/Journal label is a work that is published in a conference or journal.}
\label{fig:qa_types}
\end{figure}

\begin{figure}[h!]
\includegraphics[width=\linewidth]{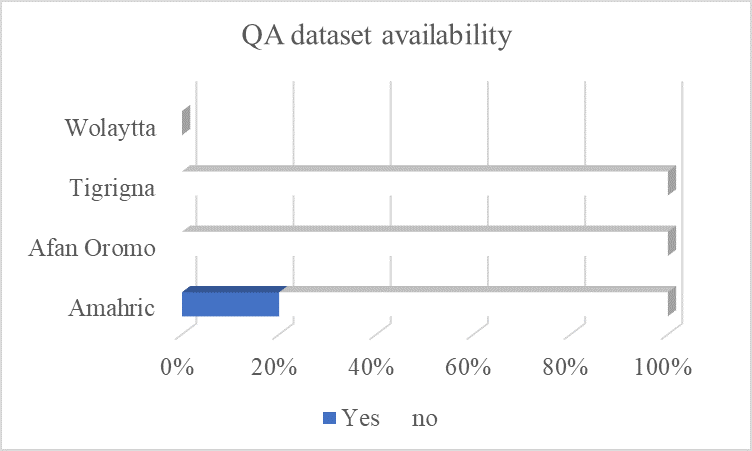}
\centering
\caption{QA dataset availability per language: as it can be seen some QC datasets are available for Amharic but not for the other languages.}
\label{fig:qa_dataset}
\end{figure}

For \textbf{Afaan Oromo},  the work by \citet{chaltu2016afaan} proposed the Afaan Oromo list, definition, and description QA system. \citet{daba2021or} improved the result of \citet{chaltu2016afaan} work for Afaan Oromo non-factoid questions. \citet{amare2016college} conducted the \textbf{Tigrinya} factoid QA system using 1200 questions. No QC or QA works have been done previously for \textbf{Wolaytta} language. Figures \ref{fig:qa_types} and \ref{fig:qa_dataset} show QC/QA publication types and dataset availability for Ethiopian languages, respectively.

%QA dataset availability

From Table \ref{fig:qa_types} and \ref{fig:qa_dataset}, we can conclude that QC and QA are less researched for Ethiopian languages, compared to the other NLP tasks. Most of the conducted works are unpublished MSc or Ph.D. theses.  Relatively, Amharic has received more attention for QA and QC tasks.

\subsection{Text Classification}

\subsubsection{Hate Speech}
Despite many works conducted on hate speech detection for resource-rich languages, low-resource languages such as Amharic, Afan Oromo, Tigrinya, and Wolaytta, are less researched. 
\begin{table*}[h!]
%\tiny
\centering
 \resizebox{\textwidth}{!}{%
\begin{tabular}{c|lrrrll}
%\toprule
\hline
 \textbf{Languages}&\textbf{Author(s)}  &\textbf{Size} & \textbf{Algorithm} &\textbf{Score} & \textbf{Dataset}  & \textbf{Model}   \\ \hline
 \multirow{5}{*}{Amharic}&\citet{mossie2018social}  & 6,120 & Word2Vec & 85.34 &No&No  \\ \cline{2-7}
 &\citet{mossie2019vulnerable} & 14,266 & CNN-GRU  & 97.85 &No &No  \\ \cline{2-7}
%&\citet{yimam2019analysis} & No  &  No &  & No &No  \\ \cline{2-7}
&\citet{abebaw2022multi} & 2,000  & MC-CNN & 74.50 & Yes & No  \\ \cline{2-7}
&\citet{bawoke2020amharic} & 30,000  & BILSTM & 90.00 &No&No  \\ \cline{2-7}
&\citet{ayele20225js} & 5,267  &RoBERTa &50.00  & Yes &Yes  \\ \hline
\multirow{3}{*}{Afaan Oromo}&\citet{ababu2022afaan}  & 12,812 & BiLSTM  & 88.00 &No &No  \\ \cline{2-7}
&\citet{defersha2021detection} & 13,600  & L-SVM   &63.00&No &No \\ \cline{2-7}
&\citet{kanessa2021automatic} & 2,780 &SVM+TF-IDF & 96.00 &No&No  \\ \hline
\multirow{1}{*}{Tigrinya}&\citet{bahre2022hate}  & 7,793 & NB+TF-IDF  &79.00 &No&No  \\ \hline
\end{tabular}
}
\caption{Summary of related works for selected Ethiopian languages in hate speech tasks, \textbf{Size} shows the number of sentences used during the experiment, \textbf{Score} shows the outperformed model results evaluated using F1 score, \textbf{Dataset} and\textbf{ Model} shows the availability of dataset and models in publicly accessible repositories, respectively.\label{tab:hatesum}}
\end{table*}
Table \ref{tab:hatesum} presents a summary of the related works in hate speech detection for selected Ethiopian languages. The table includes the name of the language, the author(s) of the paper, the size of the dataset used, the algorithm used, the score obtained, and the availability of the dataset and model in publicly accessible repositories.

For \textbf{Amharic}, five studies were conducted with different approaches. \citet{mossie2018social} used Word2Vec to detect hate speech in a dataset of 6,120 sentences and achieved an F1 score of 85.34. In another study, \citet{mossie2019vulnerable} used CNN-GRU in a dataset of 14,266 sentences and achieved an F1 score of 97.85. \citet{abebaw2022multi} used MC-CNN in a dataset of 2,000 sentences, achieving an F1 score of 74.50. \citet{bawoke2020amharic} used BILSTM on a dataset of 30,000 sentences, achieving an F1 score of 90.00. Lastly, \citet{ayele20225js} used RoBERTa on a dataset of 5,267 sentences, achieving an F1 score of 50.00. 

For \textbf{Afaan Oromo}, three studies were conducted, and none of them made their dataset or model publicly accessible. \citet{ababu2022afaan} used BiLSTM on a dataset of 12,812 sentences, achieving an F1 score of 88.00. \citet{defersha2021detection} used L-SVM on a dataset of 13,600 sentences, achieving an F1 score of 63.00. \citet{kanessa2021automatic} used SVM+TF-IDF on a dataset of 2,780 sentences, achieving an F1 score of 96.00. \citet{bahre2022hate} used NB+TF-IDF on a dataset of 7,793 sentences in the \textbf{Tigrinya} language and achieved an F1 score of 79.00. The dataset and model used in this study were not publicly accessible.
In summary, Table \ref{tab:hatesum} shows that hate speech detection in Ethiopian languages is one of the topics of research interest. However, similar to other tasks there is still a lack of publicly accessible datasets and models, which could hinder the development and evaluation of future research. It is worth noting that only two of the nine studies made their dataset and model publicly accessible. We can also see from Table \ref{tab:hatesum} that for the Wolayitta language, there is no literature found for the hate speech task. Additionally, the F1 scores obtained vary greatly among the different studies, indicating that for all tasks the results are not comparable since the datasets are different.

\subsubsection{Sentiment Analysis}

\begin{table*}[h!]
%\tiny
\centering
 \resizebox{\textwidth}{!}{%
\begin{tabular}{c|lrllll}
%\toprule
\hline
 \textbf{Languages}&\textbf{Author(s)}  &\textbf{Size} & \textbf{Algorithm} &\textbf{Score} & \textbf{Dataset}  & \textbf{Model}   \\ \hline
 \multirow{5}{*}{Amharic}&\citet{yimam2020exploring}  & 9,400 & F-Role2Vec & 58.48 &Yes &Yes  \\ \cline{2-7}
 &\citet{philemon2014machine} & 600 & Naïve Bayes  & 51.00 &No &No  \\ \cline{2-7}
%&\citet{yimam2019analysis} & No  &  No &  & No &No  \\ \cline{2-7}
&\citet{abeje2022comparative} & 2,000  & LSTM & 90.10 (accuracy) & Yes & No  \\ \cline{2-7}
&\citet{alemneh2020negation} & 30,000  & hybrid & 98.00(accuracy) &No&No  \\ \hline
\multirow{3}{*}{Afaan Oromo}&\citet{oljira2020sentiment}  & 3000 & Naive Bayes  & 93.00 &No &No  \\ \cline{2-7}
&\citet{rase2020sentiment} &  1,452  & LSTM   &87.70&No &No \\ \cline{2-7}
&\citet{wayessa2020multi} &  1,810 &SVM & 90.00 &No&No  \\ \cline{2-7}
&\citet{yadesa2020sentiment} & 341 & dictionary + contextual valance shifter  & 86.10 &No&No  \\ \hline
\multirow{1}{*}{Tigrinya}&\citet{tela2020sentiment}  & 4,000 & XLNet &81.62 &No&No  \\ \hline
\end{tabular}
}
\caption{Summary of related works for selected Ethiopian languages in sentiment analysis tasks, \textbf{Size} shows the annotated dataset used during the experiment, \textbf{Score} shows the outperformed model results evaluated using F1 score, \textbf{Dataset} and\textbf{ Model} shows the availability of dataset and models in publicly accessible repositories, respectively.\label{tab:sentsum}}
\end{table*}
Table \ref{tab:sentsum} summarizes recent studies on sentiment analysis tasks for selected Ethiopian languages, including Amharic, Afaan Oromo, and Tigrinya. The studies utilize various algorithms such as Role2Vec, Naïve Bayes, LSTM, SVM, hybrid, and XLNet. For \textbf{Amharic}, \citet{yimam2020exploring} achieved the highest F1 score of 58.48\% using Role2Vec with a dataset and a model publicly available, while \citet{abeje2022comparative} achieved the highest accuracy of 90.10\% using LSTM.
For \textbf{Afaan Oromo}, the highest accuracy of  93.00\% was achieved by \citet{oljira2020sentiment} using Naïve Bayes, while \citet{rase2020sentiment} achieved 87.70\% accuracy using LSTM. In contrast, \citet{wayessa2020multi} achieved 90.00\% accuracy using SVM. For \textbf{Tigrinya}, \citet{tela2020sentiment} achieved an F1 score of 81.62\% using XLNet with a 4000 manually labeled dataset. For \textbf{Wolaita}, similar to the hate speech task, there is no literature found for the sentiment analysis task. None of the datasets and models for Afaan Oromo, Tigrinya, and most of the works for Amharic are publicly accessible, hence results are not also comparable. This suggests that more work needs to be done in creating publicly accessible datasets and models for sentiment analysis tasks in Ethiopian languages.
In conclusion, the studies in Table \ref{tab:sentsum} indicate the potential for sentiment analysis in Ethiopian languages.  The results show that the models' performance varies depending on the algorithm, dataset, and model availability. Still, there is a need for further research to create publicly accessible datasets and models to improve the models' performance and make them available for use in different applications.

\begin{table*}[h!]
% \small
 \resizebox{\textwidth}{!}{%
\centering
\begin{tabular}{llll}
\hline
\textbf{Author(s)} & \textbf{Task} & \textbf{Language} & \textbf{dataset link} \\
%MT
\hline
\citet{gezmu2021extended} & MT & Amh-Eng & \url{http://dx.doi.org/10.24352/ub.ovgu-2018-144}\\
\hline
\citet{belay2022effect} & MT & Amh-Eng & \url{https://github.com/atnafuatx/EthioNMT-datasets}\\
\hline
\citet{abate2019english}& MT & Amh-Eng, Orm-Eng, Tir-Eng, Wal-Eng &\url{http://github.com/AAUThematic4LT/}\\
\hline
\citet{lakew2020low}& MT & Amh-Eng, Orm-Eng, Tir-Eng&\url{https://github.com/surafelml/Afro-NMT}\\
\hline
\citet{vegi-etal-2022-webcrawl}& MT & Amh-Eng, Orm-Eng&\url{https://github.com/pavanpankaj/Web-Crawl-African}\\
\hline
\citet{tedla2016tigrinya}& POS & Tir &\url{https://eng.jnlp.org/yemane/ntigcorpus}\\
\hline
%QA and NER
\citet{belay-etal-2021} & QC & Amh & \url{https://github.com/uhh-lt/amharicmodels}\\
\hline
\citet{nega2016question} & QC & Amh & \url{https://github.com/seyyaw/amharicquestionanswering}\\
\hline
\citet{tacl_a_00416} & NER & Amh & \url{https://github.com/masakhane-io/masakhane-ner}\\
\hline
\citet{jibril2022anec} & NER & Amh &  \url{https://github.com/Ebrahimc/} \\
\hline
SAY project NER dataset & NER & Amh &  \url{https://github.com/geezorg/data} \\
\hline
%SA and Hate speech 
\citet{yimam2020exploring} & SA & Amh &  \url{https://github.com/uhh-lt/ASAB} \\
\hline
\citet{ayele20225js} & hate & Amh & \url{https://github.com/uhh-lt/amharicmodels}\\
\hline
\citet{Minale2022} & hate & Amh (dataset only) & \url{https://data.mendeley.com/datasets/p74pfhz3yx/}\\
\hline
\citet{abebaw2022multi} & hate & Amh & \url{https://zenodo.org/record/5036437}\\
\hline
\citet{info12020052} & news & Tir &  \url{https://github.com/canawet/} \\
\hline
\citet{DBLP:journals/corr/abs-2103-05639} & news & Amh &  \url{https://github.com/IsraelAbebe/} \\
\hline
\citet{DBLP:journals/corr/abs-2106-13822} & text summ. & Amh, Orm, Tir&  \url{https://github.com/csebuetnlp/xl-sum} \\
\hline
\end{tabular}
}
\caption{\label{related}
Available datasets for Ethiopian languages.
}
\end{table*}

\subsubsection{News Classification and Text Summarization}
The development of an Amharic news text classification dataset is described in a publication by \citet{DBLP:journals/corr/abs-2103-05639}. The dataset consists of 50,000 sentences and is classified into six categories, including local news, sports, politics, international news, business, and entertainment. %  To promote further research and development, the dataset is publicly available, as shown in Table \ref{related}. 
\citet{info12020052} created a Tigrigna text classification dataset with manual annotation, consisting of 30k news sentences categorized into six classes, including sport, agriculture, politics, religion, education, and health. To enhance their analysis, the authors investigated the use of various word embedding techniques such as CNN, bag of words, skip-gram, and fastText. The dataset used for these experiments was made publicly available, as shown in Table \ref{related}. The work by \citet{megersa2020hierarchical} utilized a dataset collected from the Ethiopian News Agency to experiment with 8 and 20 classes, but unfortunately, both the model and datasets are not publicly available. 
%In their research, \citet{megersa2020hierarchical} utilized a dataset collected from the Ethiopian News Agency to experiment with 8 and 20 classes. To address the decrease in performance that they observed when increasing the number of classes, the authors implemented hierarchical classifiers. Unfortunately, despite efforts to locate the dataset and code necessary to replicate their experiments, it was not possible to do so. 
%
%No more news classification works are available for Ethiopian languages. This is due to researchers not giving attention to low-resource languages and the unavailability of a news corpus for Ethiopian languages.

%\subsection{Text Summarization}
\citet{DBLP:journals/corr/abs-2106-13822} created an abstractive summarization dataset for 44 different languages using BBC articles collected via crawling. % BBC's conventional approach of highlighting a concise abstract of the entire article with a brief bold paragraph at the beginning of the piece, along with other heuristics, was utilized by the authors to generate this dataset. 
The resulting dataset comprises 5461 Amharic, 4827 Tigrinya, and 5,738 Afaan Oromo samples, which can potentially be employed for various Ethiopian language-related tasks. The authors fine-tuned mt5 models using this dataset and subsequently reported the outcomes. All publicly available data and code are listed in Table \ref{related} for exploration. In general, news classification and text summarization has not yet been properly researched for Ethiopian languages.

%The same to news classification, no more text summarization works and datasets are available for Ethiopian languages.
% \section{Dataset Availability and Summary }\label{datset summary}

\section{Summary of Challenges, Opportunities and Future Directions} \label{summ}
% In this section, we discuss the challenges and future directions for each of the NLP tasks described above.
% \subsection{MT}
% For Ethiopian languages, the MT task is one of the researched areas in NLP but as \cite{belay2022effect} showed the challenges of commercially available MT systems when translating Amharic to English and other languages, this shows the MT research still needs a lot of work to improve the performance for Ethiopian languages.
\textbf{Challenges}:
% Nowadays, because of the popularity of neural network approaches like deep learning and transformer for different NLP tasks current works in MT translation also mainly focus on these state-of-the-art methods. For Ethiopian language, \cite{biadgligne2021parallel,biadgligne2022offline,belay2022effect,gezmu2021neural, chala2021crowdsourcing, gemechu2021machine,kidane2021exploration,tonja2023low} proposed NMT approaches, but the performance of this approaches highly depends on the availability of parallel dataset.
Based on the findings of the above studies, we identified the following challenges: \textbf{(i)} A scarcity of publicly available data for Ethiopian languages. As the data and resources are not mostly publicly available, researchers are going to "re-inventing the wheel" by trying to address the problem. This leaves the low-resource language research usually `in limbo`, as it is not clear if the problem is addressed or not. This further makes it difficult to train different NLP tasks for Ethiopian languages and limits the scope of NLP applications. Moreover, it is very difficult to reproduce results since the benchmark datasets are not maintained. \textbf{(ii)} A lack of resources, tools, and infrastructure for NLP research in low-resource Ethiopian languages, can make it difficult to attract funding and talented researchers to work on the problem. \textbf{(iii)} Few people are interested in NLP for low-resource Ethiopian languages. This can make it difficult to attract resources and support for NLP research in these languages.

\textbf{Opportunities:}
Here are some suggestions and ideas for the future that will help get more Ethiopian languages into NLP research: \textbf{(i)} There needs to be more work done to collect and label data in Ethiopian languages. This will require collaboration between linguists, NLP experts, and native speakers of the languages. \textbf{(ii)} As the results of the addressed NLP tasks are not comparable since the datasets are different, one big issue to address in the future is the release of benchmark datasets on which researchers can work on improving performance and developing new approaches. This will require sustained funding and collaboration among researchers. \textbf{(iii)} The development of machine translation systems for low-resource Ethiopian languages can help bridge the language gap and enable communication across different languages. \textbf{(iv)} Transfer learning techniques can be used to leverage pre-trained models in high-resource languages to improve the performance of models in low-resource languages. \textbf{(v)} The involvement of local communities and stakeholders is critical for the success of NLP research in low-resource Ethiopian languages. People in the community can give researchers and developers important information about the language and culture. \\
% \textbf{}
% \todo{
% 1. It would be interesting to see how the final work will impact research in this area beyond the preliminary findings. 
% 2. Are the findings transferable? Can  the methodology/techniques be replicated in other contexts? The authors should provide additional information on these questions.
% }
\textbf{Impact of this work and future directions}:
The results of this survey could be used to support future research initiatives in the field of NLP in Ethiopian Languages. Researchers can use the findings of the survey to identify areas that require further investigation and to develop research proposals that address the challenges and opportunities identified in the survey. This work also helps to conduct more surveys and develop a low-resource language demarcation. The demarcation helps to identify languages that need more NLP research attention. Adding more Ethiopian languages to NLP research will require researchers, linguists, and native speakers of the languages to work together, hence, at some point, these languages will be not considered low-resource languages anymore. Moreover, we point out with caution that not all the gaps and challenging problems can be instantly and readily fixed by researchers and research teams alone. Some of these problems call for sustained community cooperation as well as significant research funding from academic funding organizations. The difficulties we discussed in this paper are based on what we have learned from published research work and a quick scan of available corpora. Further studies with more comprehensive analysis, such as questionnaires directed to resource authors and users, or a more systematic inspection of the available data, can provide a deeper understanding of the causes of these problems and suggest effective solutions.

% \todo{Added this part (below this sign) to use it as a conclusion (feel free to edit or remove it)}

\section{Conclusion}
In this work, we investigated the most common NLP tasks and research  works carried out in four Ethiopian languages. We explored the main NLP research directions, progress, challenges, and opportunities for Ethiopian languages. Our findings revealed that a significant amount of research has been centered on English or Amharic-centric machine translation tasks. Despite there being a plethora of written languages in Ethiopia, only a few of them have been explored in common research studies. Additionally, we observed a low prevalence of valuable resource publications in international conference venues. The majority of works are master's theses. The publicly available datasets, models, and tools are released in a centralized GitHub repository\footnote{https://github.com/EthioNLP/Ethiopian-Language-Survey}. 
 In the future, we plan to conduct a survey on more African languages and try to come up with an NLP resource demarcation line that could help funders to prioritize research topics and languages.
\bibliography{anthology,custom}
\bibliographystyle{acl_natbib}

\appendix

\section{MT Summary} \label{appMT}
\begin{figure}[h!]
\includegraphics[width=\linewidth]{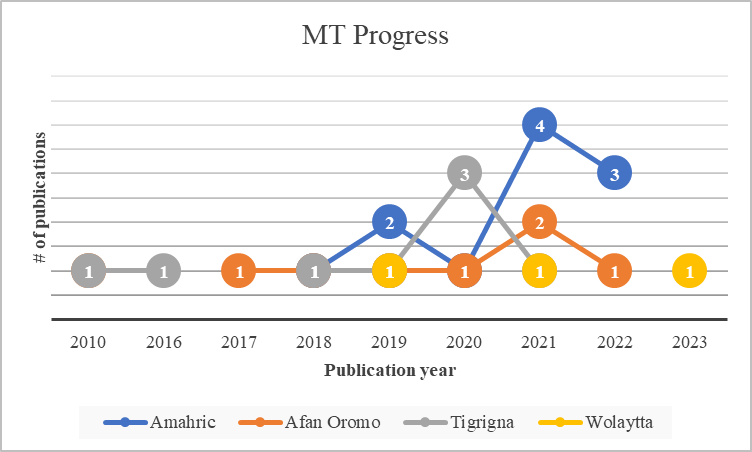} %width=8cm
\centering
\caption{MT progress per year}
\label{fig:web_app_screen}
\end{figure}

Figure \ref{fig:web_app_screen} shows the MT progress per year. As it can be seen from the figure in recent years MT research for Ethiopian languages getting attention.

\begin{figure}[h!]
\includegraphics[scale=0.5]{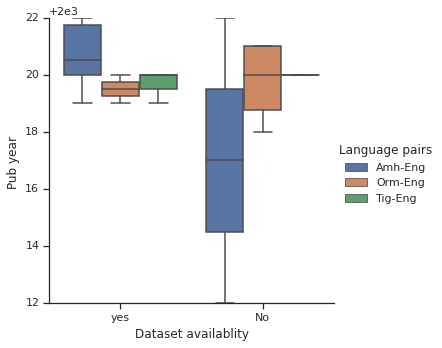} %width=8cm
\centering
\caption{(\textbf{MT=>English centeric}) Dataset availability per publication year}
\label{fig:web_app_screen1}
\end{figure}

Figure \ref{fig:web_app_screen1} shows the dataset availability per publication year. It can be noted  from the table that in recent works there are attempts to make datasets available for Ethiopian languages but this still needs more effort. 

%
% \begin{figure}[ht]
% \includegraphics[width=\linewidth]{images/Pub vs language pairs.png} %width=8cm
% \centering
% \caption{(\textbf{MT=>English centeric}) Publication year vs language pairs vs number of publication}
% \label{fig:web_app_screen2}
% \end{figure}

\begin{figure}[h!]
\includegraphics[width=\linewidth]{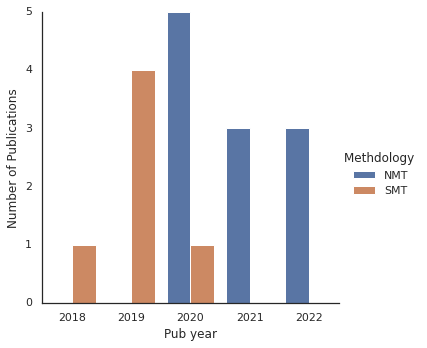} %width=8cm
\centering
\caption{(\textbf{MT=>English centeric}) Methodology per publication year }
\label{fig:web_app_screen3}
\end{figure}

Figure \ref{fig:web_app_screen3} shows the publications and methodologies used. It can be seen from the figure that before 2021 the dominant methodology used by different researchers was SMT, but in recent years different researchers have applied a neural network-based approach even if its performance depends on the availability of parallel datasets. 

\end{document}